\documentclass[pdflatex,sn-mathphys-ay]{sn-jnl}

\usepackage{graphicx}%
\usepackage{multirow}%
\usepackage{amsmath,amssymb,amsfonts}%
\usepackage{amsthm}%
\usepackage{mathrsfs}%
\usepackage[title]{appendix}%
\usepackage{xcolor}%
\usepackage{textcomp}%
\usepackage{manyfoot}%
\usepackage{booktabs}%
\usepackage{algorithm}%
\usepackage{algorithmicx}%
\usepackage{algpseudocode}%
\usepackage{listings}%
\usepackage{threeparttable}%
\usepackage{multirow}%
\usepackage{array}%
\usepackage{bbm}

\theoremstyle{thmstyleone}%
\theoremstyle{thmstyletwo}%

\theoremstyle{thmstylethree}%

\begin{document}

\title[Article Title]{Instance-Aware Pseudo-Labeling and Class-Focused Contrastive Learning for Weakly Supervised Domain Adaptive Segmentation of Electron Microscopy}


\author[1]{\fnm{Shan} \sur{Xiong}}\email{shanx\_xiong@163.com}

\author[1]{\fnm{Jiabao} \sur{Chen}}\email{23014083007@stu.hqu.edu.cn}

\author*[2]{\fnm{Ye} \sur{Wang}}\email{202261000032@jmu.ecu.cn}

\author*[1]{\fnm{Jialin} \sur{Peng}}\email{jialinpeng@hqu.edu.cn}

\affil*[1]{\orgdiv{College of Computer Science and Technology}, \orgname{Huaqiao University}, \orgaddress{\street{Jimei Rd}, \city{Xiamen}, \postcode{361021}, \state{Fujian}, \country{China}}}

\affil*[2]{\orgdiv{Department of Digital Economics}, \orgname{Jimei University}, \orgaddress{\street{Yinjiang Rd}, \city{Xiamen}, \postcode{361021}, \state{Fujian}, \country{China}}}


\abstract{Annotation-efficient segmentation of the numerous mitochondria instances from various electron microscopy (EM) images is highly valuable for biological and neuroscience research. Although unsupervised domain adaptation (UDA) methods can help mitigate domain shifts and reduce the high costs of annotating each domain, they typically have relatively low performance in practical applications. Thus, we investigate weakly supervised domain adaptation (WDA) that utilizes additional sparse point labels on the target domain, which require minimal annotation effort and minimal expert knowledge.  To take full use of the incomplete and imprecise point annotations, we introduce a multitask learning framework that jointly conducts segmentation and center detection with a novel cross-teaching mechanism and class-focused cross-domain contrastive learning. While leveraging unlabeled image regions is essential, we introduce segmentation self-training with a novel instance-aware pseudo-label (IPL) selection strategy. Unlike existing methods that typically rely on pixel-wise pseudo-label filtering, the IPL semantically selects reliable and diverse pseudo-labels with the help of the detection task. Comprehensive validations and comparisons on challenging datasets demonstrate that our method outperforms existing UDA and WDA methods, significantly narrowing the performance gap with the supervised upper bound. Furthermore, under the UDA setting, our method also achieves substantial improvements over other UDA techniques.}




\maketitle

\section{Introduction}\label{sec1}
Recent advancements in various 3D electron microscopy (EM) technologies, such as transmission electron microscopy (TEM), serial block face-scanning electron microscopy (SBF-SEM), and focused ion beam-scanning electron microscopy (FIB-SEM), enable researchers to obtain high-resolution information about intracellular structures.  Consequently, there is a pressing need for automatic segmentation of subcellular organelles \citep{aswath2023segmentation}, \textit{e.g.}, mitochondria \citep{lucchi2013learning,neikirk2023call},  from large-scale sequential cross-sectional EM images. Such segmentation can provide critical information for research in biological and neuroscience \citep{neikirk2023call,jenkins2024mitochondria}. For instance, upon reconstructing over 135,000 mitochondria, \cite{liu2022fear} found that fear conditioning not only increases the mitochondrial count but also leads to a reduction in their size.   This highlights the necessity of precise instance-level segmentation of mitochondria for meaningful insights into cellular responses.

Despite the significant advancements in deep learning methods \citep{ronneberger2015u,peng2021csnet,pan2023adaptive,zhou2021review,chen2022mask,shi2024evidential} for image segmentation, training supervised learning methods relies strongly on large-scale \textit{domain-specific training data} with \textit{high-quality fine-grained annotations}. However, collecting pixel-wise expert annotations for biomedical images \citep{qiu2024weakly,wei2020mitoem},  especially for cellular EM images that often feature numerous organelle instances, is prohibitively expensive. 
Furthermore, the performance of well-trained models can significantly decline \citep{ben2010theory,yang2024generalized}  when there is a discrepancy between the distributions of the training and testing data.
 In the case of  EM image segmentation, images acquired from diverse tissues usually exhibit considerable style/acquisition variations, along with structural/biological variations, leading to significant domain shifts. The issue is further compounded by the considerable differences in mitochondrial morphology across various cell types  \citep{pekkurnaz2022mitochondrial,jenkins2024mitochondria}. Therefore, the high cost of manual annotation, combined with the domain shifts observed in different EM datasets,  presents significant challenges for developing fully supervised models and for deploying well-trained segmentation models across varying domains.

To reduce annotation burden and to leverage well-trained models, researchers have developed various domain-adaptation methods \citep{guan2021domain,peng2020unsupervised,araslanov2021self,wu2021uncertainty,yin2023class,huang2022domain}. These methods aim to adapt a model that has been well-trained on a labeled source domain to a different target domain. Unsupervised Domain Adaptation (UDA) \citep{guan2021domain} is a widely considered setting that operates under the assumption that no labels are available in the target domain, thus completely eliminating the annotation burden. However, despite significant research efforts, a substantial performance gap persists between the UDA and supervised methods for most tasks.  This gap presents challenges when applying UDA in real-world biomedical applications \citep{ben2010theory,qiu2024weakly}, where sufficiently high segmentation performance is critically important. In practical applications, finding an effective balance between segmentation accuracy and annotation cost is essential.

\begin{figure}[t]
\centering
\includegraphics[width=0.7\textwidth]{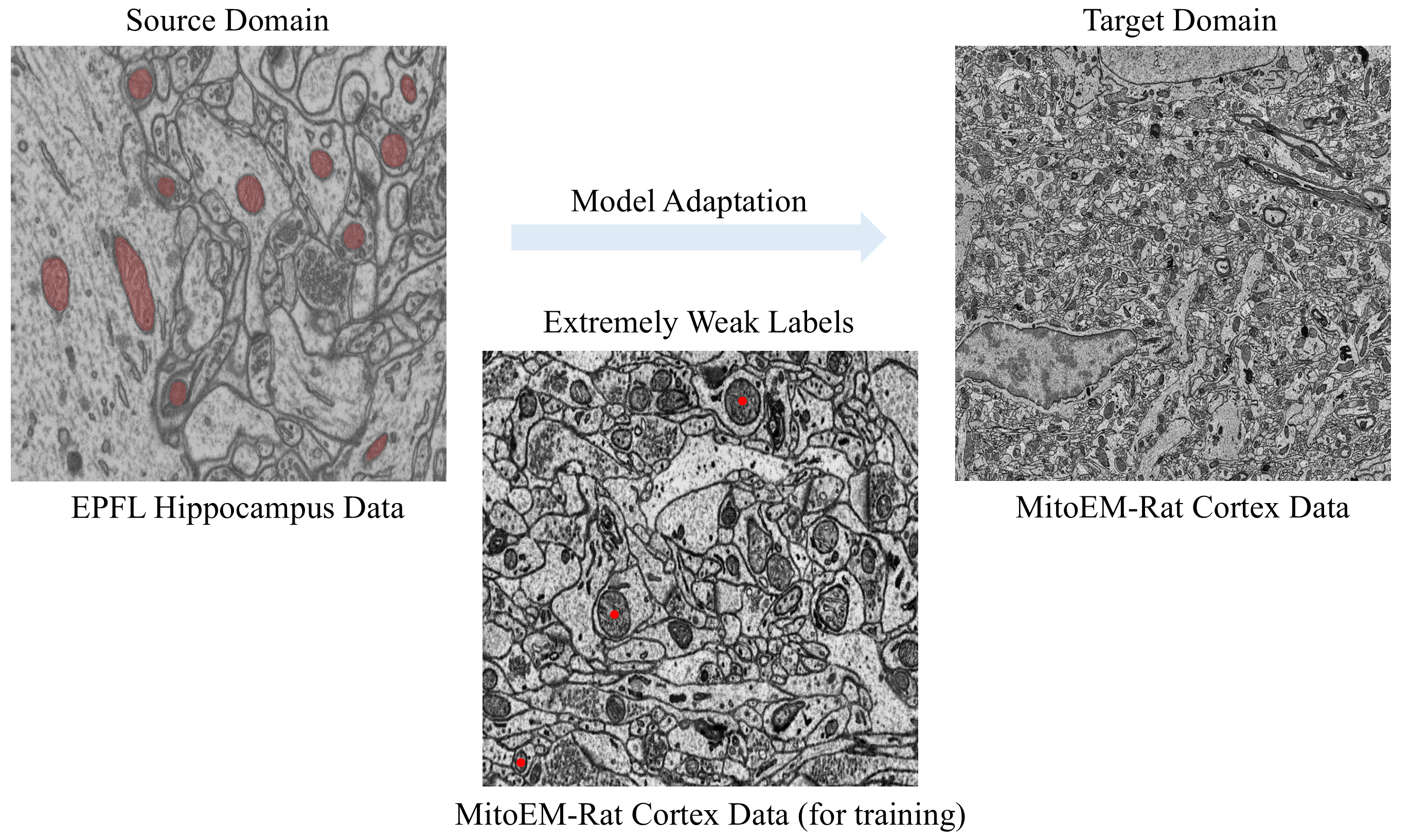}
\caption{Weakly Supervised Domain Adaptation (WDA) with sparse center points on a few object instances as weak training labels.  EM images taken from different tissues show obvious domain shifts, which pose challenges when applying models on the source domain to target domains. EPFL Hippocampus Data \citep{lucchi2011supervoxel} represent tissues from the mouse CA1 hippocampus region, while MitoEM-Rat Cortex Data \citep{wei2020mitoem} represent tissues from the primary visual cortex of an adult rat. Sparse point labels require minimal annotation efforts and reduced expert knowledge.
} \label{fig:1}
\end{figure}

To address these relatively less-explored issues, we consider weakly supervised domain adaptation (WDA), which allows extremely weak labels on the target training data,  as illustrated in Fig. \ref{fig:1}.  A significant challenge associated with full pixel-wise annotation of EM images is that it requires extensive expert knowledge to \textit{annotate all instances of the target organelle} and to \textit{delineate the exact boundaries}.
To minimize the burden of required annotations for segmenting numerous organelle instances, we utilize \textit{sparse points}—specifically, only center points on a few object instances, as introduced by \cite{qiu2022wda}—as weak labels. This form of weak annotation reduces over 95\% of the time required for pixel-wise annotations and reduces 74\% of the time required for full point annotations \citep{qiu2024weakly}. More importantly, only annotating partial center points demands significantly less expertise and knowledge, making it particularly beneficial for microscopy image annotation. It is worth noting that point annotation represents a more extreme form of the commonly used scribble and bounding box annotations. However, it is significantly easier to annotate a large number of organelle instances compared to using scribbles or bounding boxes. Additionally, annotating center points is more straightforward compared to placing effective scribbles \citep{das2023weakly} or bounding boxes \citep{lee2022bi}, which typically require more specialized domain knowledge and a deep understanding of the specific target object.

Given partial labels, \textit{self-training} \citep{amini2025self} with pseudo-labels is typically used to leverage unlabeled data. While most existing methods concentrate on selecting confident pseudo-labels to minimize error accumulation, less attention has been given to \textit{the diversity of pseudo-labels}, which refers to their representativeness. This diversity is crucial for avoiding bias in pseudo-labels. The most common strategies for quantifying the confidence of pseudo-label selection include pixel-wise thresholding based on the maximum softmax prediction scores of the model or assessing the entropy of softmax prediction scores across all classes \citep{amini2025self}. Typically,  a high confidence threshold (e.g., 0.95)  is needed to exclude more erroneous pseudo labels and reduce the negative impacts of error accumulation. However, a high threshold will result in the exclusion of correct pseudo-labels that have relatively lower confidence scores, ultimately diminishing the representativeness of the labeled regions and compromising the model’s ability to generalize. Besides the difficulty in setting thresholds, the pixel-wise pseudo-label selection for the segmentation task, which involves structural prediction, is another limitation.

This study presents a novel multitask learning-based WDA method that utilizes  \textit{cross-task teaching}, innovative \textit{semantic pseudo-labeling}, and \textit{class-focused contrastive learning}. Given the challenge of learning with incomplete and imprecise point annotations, we augment the segmentation task with an auxiliary center-point detection task and conduct cross-task teaching to enhance the accuracy of mitochondria instance localization and the selection of semantic pseudo-labels. In light of the inherent limitations associated with pixel-wise pseudo-label selection and the difficulties in establishing a reliable confidence threshold, this work introduces a simple yet effective semantic strategy termed \textit{Instance-aware Pseudo-Label} (IPL) for pseudo-label selection. This instance-aware rectification allows pseudo-labeling pixels with relatively lower prediction scores as long as they belong to a pseudo-labeled mitochondria instance. 
This approach aims to ensure the reliability of the pseudo-labels while preserving their diversity.
To enhance feature-level domain alignment and improve the compactness of category feature representations, we introduce a contrastive learning with class-focused prototypes that are different from those in previous methods \citep{das2023weakly,lee2022bi}.
We observed that the foreground class, specifically mitochondria in EM images, usually exhibits more similarity than the more complex background class.  Concretely, we compute the foreground prototype as a mix of features from both the source domain and target domain, and a mix of features from the labeled sparse points and confident pseudo-labeled regions.
Moreover, we further strengthen the detection training with supervision derived from the segmentation predictions. Different from other domain adaptation methods \citep{qiu2024weakly, Outputspace, peng2020unsupervised}, our method circumvents the need for adversarial training, which is often challenging and prone to instability. The main contribution of our study is as follows:
\begin{itemize}[]
    \item  This study develops an effective WDA method that significantly outperforms UDA methods and demonstrates comparable performance to supervised methods while requiring minimal annotation efforts and expert knowledge.
    \item To ensure the reliability of the pseudo-labels while preserving their diversity, this study proposes a simple yet effective instance-aware pseudo-labeling strategy.
    \item A class-focused contrastive learning approach has been introduced to effectively learn domain-invariant features.
\end{itemize}

\section{Related Work} \label{sec:related}
\subsection{Unsupervised Domain Adaptation}
Deep learning models trained on a specific domain often suffer from significant performance degradation when tested on datasets with shifted distributions. Although foundation models like the Segment Anything Model (SAM) \citep{kirillov2023segment} have shown strong zero-shot generalization capabilities with object-wise prompt interaction, they still face considerable challenges in biomedical applications due to persistent performance degradation under a large domain gap and the reliance on object-wise prompts. Domain adaptation techniques aim to adapt a well-trained source model to perform well on a related but different target domain, often with limited or no labeled data.  Typical approaches seek to learn a proper feature space \citep{ben2010theory}
and align the (marginal) distributions of latent representations from source and target domains by minimizing distribution discrepancy loss (e.g., maximum mean discrepancy \citep{long2015learning}), conducting adversarial alignment \citep{DANN, Outputspace,peng2020unsupervised} and contrastive alignment \citep{lee2022bi,li2023contrast}, or performing style transfer in the image space \citep{hoffman2018cycada,li2023contrast}. For example, the AdaptSegNet in \citep{Outputspace} aligns domains by conducting adversarial learning in the output space. For EM segmentation, the DAMT-Net \citep{peng2020unsupervised} conducts domain alignment in label space, feature space, and image space.  Recently, self-training or pseudo-labeling—techniques commonly used in semi-supervised learning—have also been introduced to boost the domain adaptation \citep{araslanov2021self,2018Domain}.  In this process, self-training directly estimates pseudo-labels on target domains and iteratively trains the segmentation model with both ground truth source labels and target pseudo-labels. An example of approaches in this class is the SAC \citep{araslanov2021self}, which utilizes pixel-wise pseudo-labeling within a mean-teacher framework. For sequential EM image segmentation, DA-ISC \citep{huang2022domain} takes advantage of intersection consistency to generate pseudo labels for adjacent sections. A closely-related method to ours is the CAFA \citep{yin2023class} method, which combines pixelwise pseudo-labeling with contrastive alignment, using source domain prototypes of all categories to guide feature alignment. However, due to the domain gap, especially the significant distribution shift in the background class, using the source prototypes of all classes tends to produce suboptimal results.   To adapt the SAM to medical images and integrate domain-specific medical knowledge,  the Med-SA \citep{wu2025medical} proposes to fine-tune the SAM with medical images using a lightweight  Adapter module, showing stronger performance on medical domains with proper prompt interaction.

\subsection{Weakly Supervised Domain Adaptation}
Recently,  various weak labels, such as image-level labels \citep{paul2020domain,das2023weakly}, scribbles/coarse labels \citep{das2023weakly}, points \citep{das2023weakly,zhang2024dawn}, and bounding-boxes \citep{li2022domain}, have been  explored  for cross-domain segmentation/detection. While image-level labels only provide categorical information, scribbles, bounding boxes, and points also offer information about the location and appearance of objects. Consequently, image-level labels are typically used in multi-class semantic segmentation \citep{paul2020domain,das2023weakly}.  However,  our focus is on mitochondria segmentation,  where categorical information does not significantly enhance the identification of the mitochondria instances.  \cite{das2023weakly} considered using a single point for each semantic class in multi-class semantic segmentation, whereas we consider using sparse points on a few instances to facilitate the segmentation of numerous organelle instances. They also used the features derived from weak labels as anchors/prototypes to guide contrastive feature learning. However, with point annotations, the annotated areas are quite
limited, and the weak-label-based anchor may be misleading, especially in complex biomedical image segmentation tasks.  A closely-related work is the WDA-Net \citep{qiu2024weakly}, which introduces sparse points as weak labels and employs an adversarial domain alignment. They also use self-training, but with a pixel-wise entropy-based pseudo-label estimation strategy.

\section{Methodology} \label{sec:method}
\subsection{Overview}
\textbf{Problem setting}.
In the  WDA setup,   we are given densely labeled source data $\mathcal{D}^s=\{(x^s,y^s)\}$, where $x^s \in \mathbb{R}^{H\times W} $ is a source image and $y^s\in \mathbb{R}^{H\times W}$ is its corresponding label image, and a weakly-labeled target dataset $\mathcal {D}^t=\{(x^t,\bar{c}^t)\}$, where $\bar{c}^t\in\{0,1\}^{H\times W}$ is the point label only taking 1 on the sparsely annotated points. Additionally, we denote the corresponding point label of $y^s$ as $c^s\in\{0,1\}^{H\times W}$, which only takes 1 on the center point of each object instance in $y^s$. We further denote the corresponding density map of $c^s$  as $d^s$ that can be obtained through the convolution of the point label with a Gaussian kernel $k_\sigma$, i.e., $d^s=k_\sigma*c^s$, in which $\sigma$ (e.g., 11) is the kernel bandwidth.
Our goal is to adapt the segmentation model trained on the source domain to the target domain.
\begin{figure*}[t]
\centering
\includegraphics[width=0.9\textwidth]{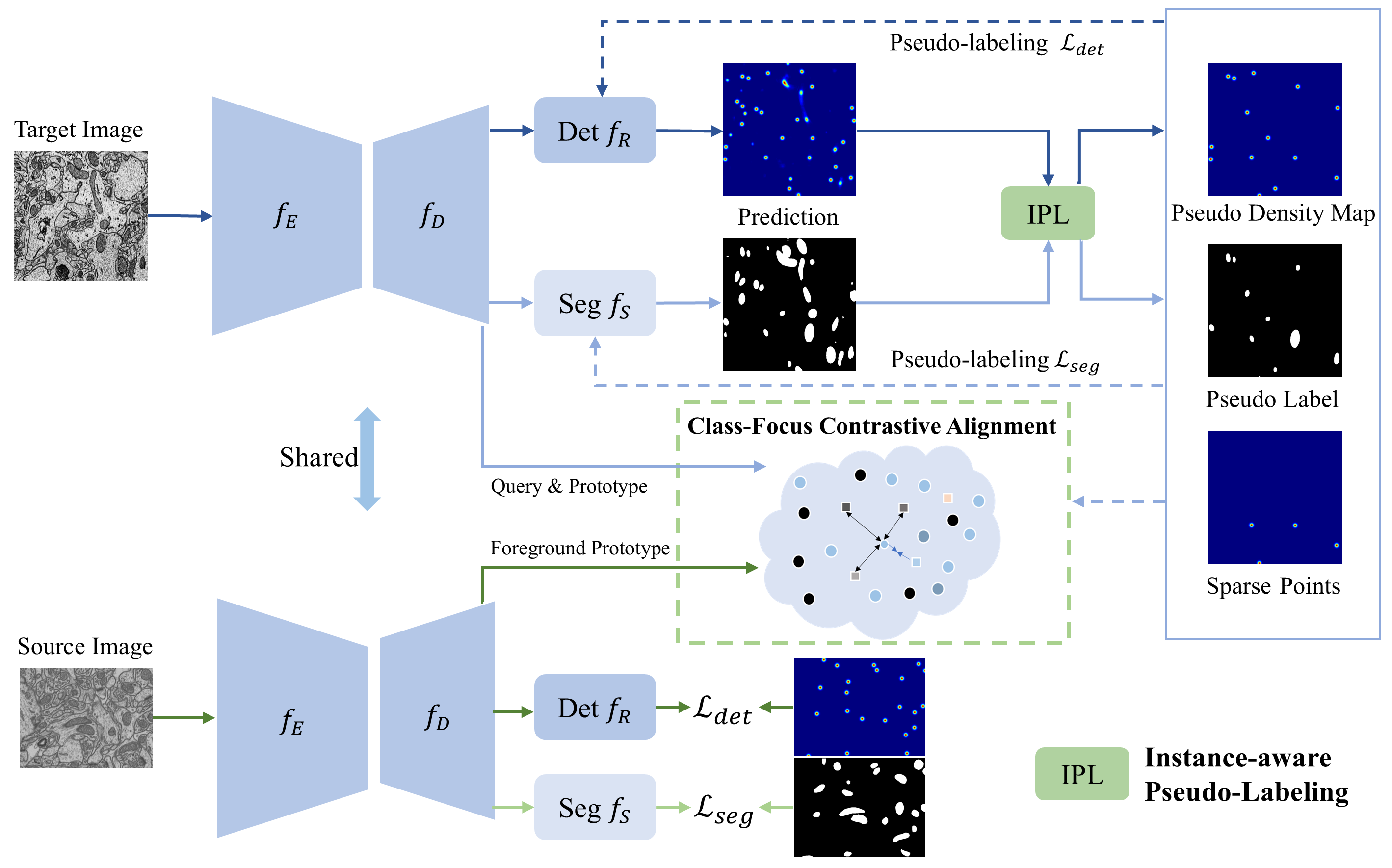}
\caption{Overview of the proposed method for weakly-supervised cross-domain adaptation. Under a multitask learning framework, an auxiliary center detection task is utilized to achieve instance-aware pseudo-label selection for the self-training on the segmentation head. class-focused conservative learning through pixel-to-prototype alignment is introduced to conduct feature-level domain alignment and improve the compactness of category feature representations.
  } \label{fig:2}
\end{figure*}

\textbf{Model overview}.
An overview of our model is demonstrated in Fig. \ref{fig:2}, in which we conduct multitask learning with a cross-task teaching mechanism. Our model takes an encoder-decoder architecture $f_D\circ f_E$ with a segmentation head $f_S$ and a regression-based detection head $f_R$, where $f_E$ is the shared feature encoder and $f_D$ is the shared base decoder. The segmentation sub-network $F_{S}=f_S\circ f_D\circ f_E$  predicts the probabilistic segmentation map, and the detection sub-network $F_{D}=f_R\circ f_D\circ f_E$ predicts the density map of the center points of target object instances. $F_{S}$ and $F_{D}$  are jointly pre-trained on the labeled source domain. At the model adapting stage, the training of both $F_S$ and $F_D$ is supervised with source labels, target sparse point labels, and target pseudo labels. Throughout the training process, the output of the detection sub-network $F_D$ is used to guide the selection of reliable and diverse pseudo-labels for the segmentation.
Additionally, the segmentation sub-network $F_S$ is also guided by a contrastive alignment loss, which significantly enhances domain-invariant feature learning. 

\subsection{
Auxiliary detection task}
Compared to the complicated dense segmentation task, detecting instance centers is much easier, particularly when using only sparse point annotations. Furthermore, the outputs from the detection can be effectively leveraged to substantially enhance the segmentation of mitochondrial instances. This insight leads us to investigate a unified learning framework that encompasses both segmentation and detection tasks. For center-point detection, we adopt the density map regression strategy \citep{zhang2016single}. Notably, although the annotated sparse points provide minimal additional information to the dense segmentation task, they provide much richer information to the center-point detection task.

\textbf{Detection pseudo-label learning}. To adapt the source detection model to the target domain with the sparse point annotations, we utilize the self-training strategy \citep{lee2013pseudo}. This involves  iteratively train our detection model using a combination of ground-truth sparse point labels and the progressively refined pseudo-labels.  A key aspect  of this self-training process is the effective selection of pseudo-labels, which is critical for improving model performance.
\begin{equation}\label{eq:d}
\mathcal{L}_{det}=\frac{1}{|\mathcal{D}^s|}\sum_{x^s} L_{2}(F_{D}(x^s),d^s)+\frac{1}{|\mathcal{D}^t|}\sum_{x^t} L_{2}(F_{D}(x^t),\hat{d}^t)
\end{equation}
where $L_{2}(\cdot,\cdot)$ is the standard least square loss, $\hat{d}^t$ is the density map of the union set of ground-truth sparse points and the estimated pseudo-label points.
More specifically, in each self-training iteration, we first estimate the total number of mitochondria instances by summing the predicted density map. Then, we select a proportion of $\beta$\% (e.g., 20\%) more points that have the highest responses in the predicted density map as the additional pseudo-label points.  Given a density map,  candidate points with the highest responses are local maxima points and can be obtained with standard Non-Maxima Suppression (NMS) \citep{zhang2016single}.  To suppress noises in the predicted density map, we ensemble the predictions of 5 differently augmented versions (e.g., flip, rotate, etc.) of the input. Ground-truth sparse points are always used for supervision.

Thus, in each self-training iteration, we integrate the newly selected points with previously selected points as updated pseudo-labels and then construct new density maps to supervise the model training. For the target data, the $L_{2}$ loss in $\mathcal{L}_{det}$ is calculated exclusively on positive regions of $\hat{d}^t$ and confident background regions inferred from the density map in the preceding self-training iteration. Our experimental setup involves training the detection model over four iterations, utilizing up to 80\% of the center points. Additionally, by maintaining a shared architecture across most layers, the detection network $F_D$  implicitly enhances the segmentation network $F_S$, enabling more accurate instance recognition.

\subsection{Segmentation via instance-aware pseudo-labeling}
Due to the extremely limited number of labeled pixels in the target training data, as illustrated in Fig. \ref{fig:1}, we train the segmentation sub-network  $F_S$  under the framework of self-training, which is guided by the detection predictions. To supervise the model training, we utilize a combination of dense labels from the source, point labels from the target, and generated pseudo-labels for the target.
\begin{equation}\label{eq:1}
\mathcal{L}_{seg}=\frac{1}{|\mathcal{D}^s|}\sum_{x^s} L_{CE}(p(x^s),y^s)+\frac{1}{|\mathcal{D}^t|}\sum_{x^t} L_{CE}(p(x^t),\hat{y}^t)
\end{equation}
where $L_{CE}(\cdot,\cdot)$ is the standard cross-entropy loss, $p(x)=F_S(x)$ is the predicted segmentation,
$\hat{{y}}^t$ is a fusion of the estimated pseudo-labels and ground truth point annotations.

\textbf{Instance-aware pseudo-label selection}. Instead of using a single threshold for filtering pixel-wise pseudo-labels, we select confident pseudo-labels on an instance-wise basis to preserve the diversity of pseudo-labeled regions.
Specifically, we propose to leverage the output of the detection head $F_D$  to guide pseudo-label selection during the segmentation. In each self-training iteration, we first estimate the total number of mitochondrial instances by summing the predicted density map.  Then, we extract a specified proportion (for example,  50\%) of the most confident local maxima from the density map. We refer to this set of selected points as  $\Omega$. Ground-truth sparse points are incorporated as confident points as well. Subsequently, we identify the connected regions associated with these confident points in the current binary segmentation output to serve as our pseudo labels for mitochondria instances.

 \begin{equation}
 \hat{y}^t={\rm IPL}(\Omega,{\rm TH}(p(x^t)))
 \end{equation}
 where $\Omega$ is the selected confident centers, ${\rm TH}(p(x^t))$ denotes the binarization of the probabilistic segmentation prediction of $x^t$ at a predefined threshold (for instance, 0.7 in our experiments), ${\rm IPL}(\cdot,\cdot)$ signifies the instance-aware pseudo-label selection that is achieved through selecting the connected regions associated with  $\Omega$ from ${\rm TH}(p(x^t))$.  Here, we refer to these selected connected regions as \textit{pseudo-label instances}. In our experiment setup, we conduct three iterative rounds of self-training. During the first iteration, we restrict ourselves to using only the ground-truth sparse points as the confident points. In the subsequent second and third rounds, we expand our selection to 50\% and 95\% of the most confident points, respectively, to select pseudo-label instances.  To further enhance segmentation accuracy,  we also use the detection predictions to filter out false positives.

\subsection{Class-focused contrastive alignment}
\textbf{Contrastive feature alignment}. To enhance feature-level domain alignment and improve the compactness of category feature representations,  we augment our model with class-focused pixel-to-prototype contrastive learning. Unlike pixel-to-pixel contrast, pixel-to-prototype contrast with well-defined prototypes can yield greater robustness and computational efficiency. In this study, we exploit multiple prototypes for each class. Let $z^t=f_D\circ f_E(x^t)$ denote the feature maps derived from the target domain input $x^t$. For each class $c$,  we extract a set of pixel-wise queries
$\mathbb{Q}^c=\{z_{ic}^t\}_{i=1}^{N}$ solely from the target domain,  and  a set of class prototypes
$\mathbb{P}^c=\{\mu_k^c\}_{k=1}^{K_c}$ from the source and target domains as well as their (weighted) mean prototype $\bar{\mu}^c$.  Notably, we have observed that the foreground class in different EM images generally exhibits higher intra-class similarity compared to the more complex and varied background class. As a result, we utilize foreground prototypes estimated from features from both the source and target domains, while the background prototypes are derived exclusively from features in the target domain.

To achieve inter- and intra- domain alignment, we utilize a contrastive loss that aligns class-specific queries with the mean prototype of each class while ensuring that they are dissimilar. This strategy not only facilitates feature discrimination but also fosters effective domain adaptation across different categories through our class-focused contrastive alignment.
\begin{equation}
\mathcal{L}_{contra}=-\frac{1}{NC}\sum_{c=0}^{C-1} \sum_{i=1}^{N}
\log\frac{\exp(z_{ic}^{t}\cdot\bar{\mu}^c/\tau)}{\exp(z_{ic}^t\cdot\bar{\mu}^c/\tau)+ \sum_{l\neq c}\sum_{k=1}^{K_l}\exp(z_{ic}^t\cdot\mu_k^l/\tau)}
\end{equation}
where the temperature $\tau$ is to control the softness of the softmax function, and $C$ is the number of categories. For binary segmentation, $C$=2.

\textbf{Query samples from target domain}.  In this study, we focus solely on sampling queries from the target domain.
For each class $c$, our query set is composed of both hard samples and easy samples, represented as
$\mathbb{Q}^c=\mathbb{Q}_{hard}^c\cup\mathbb{Q}_{easy}^c$.

\textit{Easy queries} are features of the pixels that not only take the ground-truth label $c$ or pseudo-label  $c$, but also demonstrate a sufficiently high level of confidence. \begin{equation}\mathbb{Q}_{easy}^c=\{z_{i}^t|\hat{y}_i^t =c,p^t_i>\delta_e\}\end{equation}
where $p^t_i$ refers to $p(x^t)$ at pixel $i$.

\textit{Hard queries} are features of the pixels whose labels/pseudo-labels are  $c$ but show relatively low confidence.
\begin{equation}\mathbb{Q}_{hard}^c=\{z_{i}^t|\hat{y}_i^t =c,p^t_i<\delta_h\}\end{equation}
In our experiments, we set $\delta_e=0.95$ and $\delta_h=0.80$. For each class, to reduce computation complexity, we only sample $N$=1024 queries, which contain 512 hard samples and 512 easy samples, in each batch.

\textbf{Class-focused prototypes}. Since the foreground class, specifically mitochondria in EM images, usually exhibits higher similarity over different domains than the complicated background class, we introduce significantly different prototypes from those in previous methods and focus on the inter-domain alignment of the foreground class.

For the foreground class ($c$=1), the set of foreground prototypes $\mathbb{P}^1$ contains three types of prototypes: 1) the source prototype $\mu_s^{1}$ derived from the features of foreground class pixels in the source images, 2) the target label prototype $\mu_t^{1}$ estimated using the ground-truth point labels of target images, 3) the target pseudo-label prototype $\mu_{pl}^{1}$ estimated using pseudo-labels assigned to the target images. More specifically, both the source prototypes $\mu_s^{1}$ and the target label prototype $\mu_t^{1}$  are calculated by averaging the features of the correctly labeled pixels.
\begin{equation}
\mu_k^{1}=\frac{\sum_{i}z_{i}^{k}\mathbbm{1}[y_{i}^{k}=1]\mathbbm{1}[\hat{p}_{i}^{k}>\delta_e]}{\sum_{i}\mathbbm{1}[y_{i}^{k}=1]\mathbbm{1}[\hat{p}_{i}^{k}>\delta_e]},
\mu_{pl}^{1}=\frac{\sum_{i}z_{i}^{t}\mathbbm{1}[\hat{y}_{i}^{t}=1]}{\sum_{i}\mathbbm{1}[\hat{y}_{i}^{t}=1]}
\end{equation}
where $k\in\{s,t\}$, and $\mathbbm{1}[\cdot]$ is the indicator function.  While the source foreground prototype $\mu_s^{1}$ and $\mu_t^{1}$ guide cross-domain feature alignment, $\mu_t^{1}$ and $\mu_{pl}^{1}$ guide compact feature learning, which is also crucial for improving performance in the target domain.

For the background class ($c$=0), we have two types of prototypes: 1)  target pseudo-label prototype $\mu_{pl}^{0}$ estimated using pseudo-labels of the target images, 2) 2  clusters of features at pseudo-labeled regions of target images, which are computed using the K-means clustering algorithm and updated every 1000 iterations in the experiments. Note that prototypes of background features of the source domain are not considered in our method due to the large distribution shift for the background class.
\subsection{Training objectives}
An illustration of our whole model is in Fig. \ref{fig:2}.  The overall loss for model training  is given as,
\begin{equation}
\mathcal{L}_{obj}=\mathcal{L}_{seg}+\lambda_1\mathcal{L}_{det}+\lambda_2\mathcal{L}_{contra}
\end{equation}
in which $\lambda_1$ and $\lambda_2$ are non-negative trade-off parameters. During the testing stage, no user annotation is needed.

\section{Experiments} \label{sec:result}
\subsection{Datasets and metrics}
We evaluate our models using three challenging datasets.

\textbf{EPFL Hippocampus Data (EPFL)}. This dataset was scanned with a focused ion beam scanning electron microscope (FIB-SEM) \citep{lucchi2013learning} in a resolution of 5$\times$5$\times$5 $nm^3$, representing tissues from the mouse CA1 hippocampus region. This dataset was split into two subsets, each of which contains 165 image slices of size 768$\times$1024, for training and testing, respectively.

\textbf{MitoEM-R Cortex Data (R)}.
 The MitoEM Dataset \citep{wei2020mitoem} constrains two subsets. The MitoEM-R subset was scanned using a multi-beam scanning electron microscope (SEM) at a resolution of 8$\times$8$\times$30 $nm^3$, representing  Layer II/III in the primary visual cortex of an adult rat. This dataset was split into a subset of size 400$\times$4096$\times$4096 and a subset of size 100$\times$4096$\times$4096 for training and testing, respectively. To train our model, we just randomly selected 40 slices from the training set as actual training data. 

\textbf{MitoEM-H Temporal Lobe Data (H)}. MitoEM-H is a subset of the MitoEM Dataset \citep{wei2020mitoem}. This dataset was taken from Layer II in the human temporal lobe at a resolution of 8$\times$8$\times$30 $nm^3$. This dataset was also split into a subset of size 400$\times$4096$\times$4096 and a subset of size 100$\times$4096$\times$4096 for training and testing, respectively.  To train our model, we just randomly selected 40 slices from the training set as actual training data.

The EPFL dataset is the smallest of the three datasets. In contrast, the  MitoEM-R and MitoEM-H have 14.4k and 24.5k mitochondria instances, respectively, which is over 500 $\times$
more than the EPFL dataset. Additionally, the MitoEM datasets exhibit a significantly greater diversity of mitochondrial shapes and densities.

\textbf{Metrics}. To evaluate the segmentation performance, we use both Dice similarity coefficient (Dice), Aggregated Jaccard-index (AJI) \citep{kumar2017dataset}, and Panoptic Quality (PQ) \citep{kirillov2019panoptic} as the measures. Notably, AJI and PQ are instance-level measures.

Let $P$ and $G$ represent the predicted segmentation and the ground truth segmentation, respectively. The Dice coefficient can be mathematically defined as follows,
\begin{equation}
{\rm Dice}(P,G) = \frac{2 |P\cap G| }{ |P| +|G|}\times 100\%.
\end{equation}

Let  $G^j$ represent the $j \rm th$ instance (i.e., mitochondrion) in the set $G$,  with a total of
$M$ instances. Similarly, $P^j$ denotes the $j \rm th$ instance in the set $P$. The measure AJI is mathematically defined as follows,
\begin{equation}
{\rm AJI}(P,G) = \frac{\sum_{j=1}^{M} |P^j\cap G^{j^*}| }{\sum_{j=1}^{M} |P^j\cup G^{j^*}| +\sum_{i\in {\rm FP}} |P^{i}|}\times 100\%,
\end{equation}
where $j^*$ refers to the index of the matched instance in  $P$ with maximal overlapping with $G^j$; false positive (FP) refers to the set of instances in $P$ that lack corresponding ground truth mitochondria in $G$.

 The PQ is essentially a composite measure that evaluates detection accuracy and segmentation quality based on true positives (TP)\citep{kirillov2019panoptic}. Let FN represent the set of false negatives.  The PQ is mathematically defined as follows,
\begin{equation}
{\rm PQ}(P,G) = \frac{\sum_{j\in {\rm TP}} {\rm IOU}(P^j, G^{j^*}) }{{2\rm |TP|}+|{\rm FP}|+|{\rm FN}|}\times 100\%,
\end{equation}
where
${\rm IOU(P^j, G^{j^*})}= |P^j \cap G^{j^*}|/|P^j\cup G^{j^*}|$.

\subsection{Implementation details}
For fair comparison with existing methods, we use the same network architecture as the WDA-Net \citep{qiu2024weakly}. We implemented our models using Python (version 3.11) and the Pytorch framework (version 12.1)  and trained models with one NVIDIA GeForce RTX 3060 GPU with 12 GB of memory. The widely-used SGD optimizer was used for model training with an initial learning rate of $5\times 10^{-5}$ and a batch size of 1. The training process ran for a maximum of 100,000 iterations, using randomly cropped input patches of size 512$\times$512. We used a polynomial decay of power 0.9 to control the learning rate decay. The same data augmentations as the WDA-Net were also applied here. We set $\lambda_1$=$1\times 10^{-2}$ and $\lambda_2$=$5\times 10^{-3}$.  The code is available at \url{https://github.com/x-coral/WSDA}.

\begin{table*}[t]
\caption{Quantitative Comparison on three domain adaptation tasks. By default, we use 15\% of the center points as the weak annotation.  For the R and H datasets, we use only 1/10 of the training set for model training. The `Interact' means conducting inference with point prompts on all instances. The Med-SA was finetuned on the fully labeled source domain.}
\centering
 \setlength{\tabcolsep}{0.5mm}
 \label{tab:1}
\begin{tabular}{lccccccccccc}
\toprule
\multirow{2}{*}{Methods} &\multicolumn{3}{c}{EPFL $\rightarrow$ R} &~ &\multicolumn{3}{c}{R $\rightarrow$H} & ~
&\multicolumn{3}{c}{H $\rightarrow$R}\\
\cmidrule{2-4}\cmidrule{6-8}\cmidrule{10-12}
& Dice  & AJI  & PQ &~  & Dice & AJI  & PQ  & ~
& Dice& AJI&PQ  \\
\midrule
NoAdapt: Methods without Adaption &&&&&&&&&&&\\
\midrule
Our source model                &67.0& 46.0& 30.5
&~ &73.8& 57.1& 43.0 & ~
& 81.5& 65.2&58.0\\
SAM \citep{kirillov2023segment}     &32.0  &14.3  &30.0 &~ &20.8  &11.4 &18.7 &~ &32.0  &14.3  &30.0\\  
SAM (Interact) &40.6  &1.2  &26.2 & ~ &40.3 & 4.6  &26.6 &~  &40.6  &1.2  &26.2\\ 
Med-SA  \citep{wu2025medical} &36.4 &20.6 &9.7 & ~&72.4 &55.0 &33.4&~&75.5 &56.6 &27.5\\
Med-SA (Interact) &61.6 &40.1& 22.0 &~ &83.8 &68.1& 59.0&~ &86.2  &70.2  &59.9 \\
\midrule
UDA Methods&&&&&&&&&&&\\
\midrule
AdaptSegNet \citep{Outputspace}
& 77.1& 59.2& 50.1
&~ & 84.7& 71.2&58.1 & ~
& 86.5& 72.8&61.6\\
DAMT-Net \citep{peng2020unsupervised}        & 83.4& 67.9& 60.2&~ & 85.4& 72.3& 63.7 & ~
& 88.7& 76.3&61.8\\
SAC \citep{araslanov2021self}        &84.4&70.0&55.4
&~ & 77.9&60.7&37.6 & ~
& 83.1& 67.4&40.3\\
UALR \citep{wu2021uncertainty} & 76.9& 59.7&33.1&~ & 83.8& 69.7&60.0 & ~
& 86.3& 71.6&53.7\\
DA-ISC (2.5D) \citep{huang2022domain}     & 78.5& 61.5& 51.8
& ~ & 85.6& 72.7&63.8 & ~
& 88.6& 75.7&65.8\\
CAFA (2.5D) \citep{yin2023class}    & -& -&-&~ & 86.6& -&- & ~
& 89.2& -&- \\
WDA-Net (0\%) \citep{qiu2024weakly} &86.1 &71.5 &61.2 &~ &85.5 &72.3 &60.6  & ~
& 88.2& 74.5&59.0\\
Ours (0\%) & 90.2& 77.0& 72.0& &87.0&75.1&65.8 &
& 92.1& 81.6&74.7\\
\midrule
WDA Methods&&&&&&&&&&&\\
\midrule
WDA-Net (15\%)   & 90.1& 77.5& 73.1
&~ & 88.7& 77.6&67.8 & ~
& 91.7& 80.7&74.0\\
Ours (15\%)  &92.6 & 81.7& 75.6
&~ & 89.2 & 78.3 & 70.2 & ~
& 93.2& 83.6&75.9\\
\midrule
Oracle &&&&&&&&&&&\\
\midrule
 Supervised model     & 93.6& 83.8& 77.1&~ & 92.5& 84.1& 73.3 & ~ & 93.6& 83.8&77.1\\
\bottomrule
\end{tabular}
\footnotetext[1]{The Med-SA was finetuned using the fully labeled source data.}
\end{table*}

\subsection{Comparison with the State-of-the-Art methods}
We compare our method, including its UDA version, with four categories of methods.
First, we compare our methods to the State-of-the-Art (SOTA) \textbf{UDA methods}: AdaptSegNet \citep{Outputspace}, DAMT-Net \citep{peng2020unsupervised}, SAC \citep{araslanov2021self}, UALR \citep{wu2021uncertainty}, DA-ISC \citep{huang2022domain}, CAFA \citep{yin2023class}, WDA-Net (UDA)\citep{qiu2024weakly}. Second, we compare our methods with the SOTA \textbf{WDA method}, WDA-Net (15\%)\citep{qiu2024weakly}. Among the UDA methods, DA-ISC \citep{huang2022domain}, CAFA \citep{yin2023class} are 2.5D methods that take multiple slices as input, while others, including our model, simply conduct 2D segmentation. Third, we also evaluate our methods against typical \textbf{Foundation models}, including SAM \citep{kirillov2023segment}, Med-SA \citep{wu2025medical}, and their interactive versions, i.e., SAM (Interact), and Med-SA (Interact), which take center-points of all mitochondria instances as user prompts. Note that SAM was trained on billion-scale datasets, while the Med-SA is an adapted version of SAM that specifically utilizes large medical datasets and the EM data of the source domain. Fourth,  we assess our methods against the \textbf{Supervised model} that is trained with full pixel-wise labels on the target domain.

  \begin{figure}
\centering
\includegraphics[width=1\textwidth]{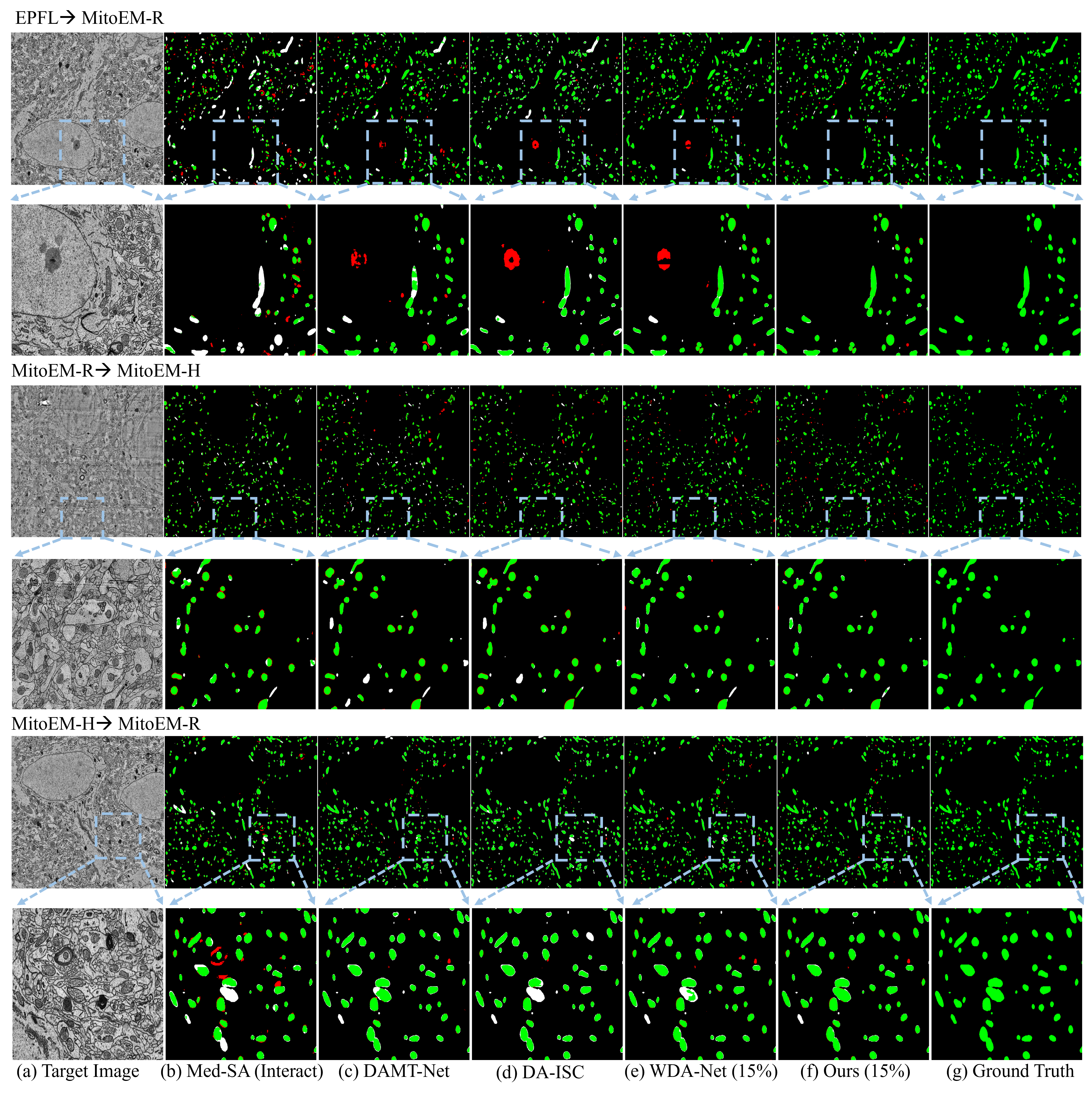}
\caption{Qualitative comparison results. Green:  true positives; Red:  false positives;  White: false negatives. For better visualization, we zoom in on the images in the blue boxes. Note that Med-SA (Interact) uses full point prompts during inference, while other methods require no prompts.   } \label{fig:3}
\end{figure}

Table \ref{tab:1} presents the comparison results across three domain adaptation scenarios:   EPFL$\rightarrow$R, R$\rightarrow$H, and H$\rightarrow$R. Notably, when compared to the fully supervised upper bound, our approach—utilizing only 15\% center point annotations—achieves a very small performance gap of 1.0\% in Dice for the EPFL$\rightarrow$R task,  3.3\% for R$\rightarrow$H, and 0.4\% for  H$\rightarrow$R. These results indicate that leveraging sparse point weak labels, which require minimal annotation effort, can effectively narrow the performance gap between supervised and domain adaptation methods. Additionally,  our method significantly surpasses the current SOTA UDA methods.  In particular, our model shows substantially enhanced performance over UDA methods when evaluated using instance-level metrics such as AJI and PQ.
Moreover, the UDA version of our approach also achieves the highest performance among the UDA methods, even demonstrating results that are competitive with those of WDA-Net (15\%).

Moreover,  using the same set of  15\% center point annotations, our model consistently outperforms WDA-Net (15\%)  across all metrics for each of the three tasks. Additionally, our model (15\%) also outperforms its UDA version by 2.4\% in Dice for  EPFL$\rightarrow$R,  2.2\% for R$\rightarrow$H, and 1.1\% for  H$\rightarrow$R.
A more detailed comparison of our method with WDA-Net, under varying amounts of point annotations, is presented in Table \ref{tab:5}  and will be discussed in the following sections.

Besides, our model significantly outperforms the SAM model and its interactive version, which uses additional point prompts during the testing stage. The poor performance of SAM highlights the large domain gap between the EM data and the training data of the SAM. Although Med-SA, the adapted version of SAM using medical data and the source EM data, shows improved performance, its performance is still significantly lower than ours. This is especially evident when fine-tuned with the small EPFL dataset.   Even when using all center points, Med-SA (Interact) still shows much lower performance than our approach.

\textbf{Visual comparison}.
Figure \ref{fig:3} provides a visual comparison of  our method with Med-SA (Interact), DAMT-Net, DA-ISC, and WDA-Net (15\%) on all three tasks. As illustrated in Fig. \ref{fig:3}, our method shows significantly reduced false positives (indicated in Red) and false negatives (indicated in White). As shown in the first row of the figures, our method demonstrates a stronger ability to successfully segment large mitochondria of irregular shapes.

\begin{table*}[t]
\centering
\caption{Ablation study of our model (15\%) on EPFL Hippocampus$\rightarrow$MitoEM-R.}
\label{tab:2}
\begin{threeparttable}
\begin{tabular}{lccccc}\\
\toprule
\multirow{2}{*}{Setting}  &\multirow{2}{3cm}{\centering Detection Pseudo-Labeling} &\multirow{2}{3cm}{\centering Segmentation Pseudo-Labeling} &\multirow{2}{3.5cm}{\centering  Class-focused \\Contrastive Learning}    &Dice\\
&~&~&~& (\%)      \\
\midrule
I   & ~          &\checkmark \tnote{$\dag$}          & ~ &84.2    \\
II  &\checkmark  & ~          & ~          &88.8    \\
III     &\checkmark &\checkmark \tnote{*}        & ~&91.1    \\
Full   &\checkmark &\checkmark \tnote{*}  &\checkmark    &\textbf{92.6}  \\
\bottomrule
\end{tabular}
\begin{tablenotes}
        \footnotesize
        \item[$\dag$] Threshold-based Pseudo-Labeling
        \item[*] Instance-aware Pseudo-Labeling
\end{tablenotes}
\end{threeparttable}
\end{table*}

\begin{figure}[t]
\centering
\includegraphics[width=0.7\textwidth]{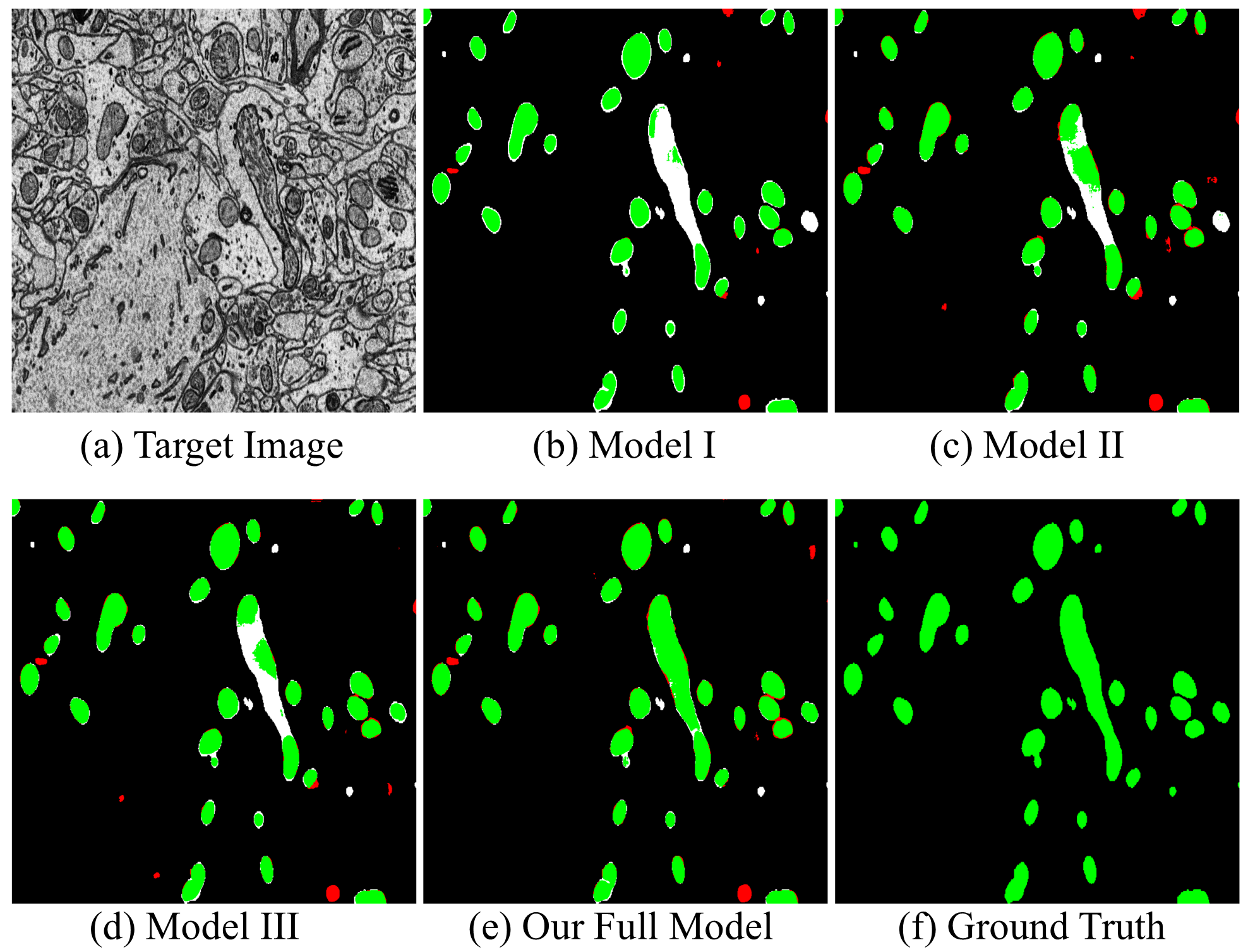}
\caption{Visual comparison of the ablated versions of our proposed approach.} \label{fig:4}
\end{figure}

\subsection{Model analysis} \label{subsec: 23}
\textbf{Ablation study of key components}.  In this section, we validate the effectiveness of our three key components:  1) auxiliary detection with pseudo-labeling, 2)  instance-aware pseudo-labeling for segmentation, and 3) class-focused
contrastive learning. By comparing our full model with Model III in Table \ref{tab:2},  we observed a performance drop of 1.5\% in the Dice coefficient when class-focused contrastive learning was removed. By further removing instance-aware pseudo-labeling, Model II resulted in a significant performance drop of 2.3\% in Dice, while only using Pseudo-Labeling-based segmentation further led to a significant performance drop of 3.8\% in Dice. Our model achieves the best performance when all three components are integrated,  demonstrating the effectiveness of our proposed approach. Figure \ref{fig:4} visually compares our model with its ablated versions, further indicating the effectiveness of the proposed components in our model.

\textbf{Comparison of  pseudo-labeling strategies}.
Table \ref{tab:3} presents a comparison of the proposed instance-aware pseudo-labeling with the widely used threshold-based method, which selects confident pseudo-labels based on predetermined thresholds, and the entropy-based strategy outlined in \citep{qiu2024weakly}. Our instance-aware strategy demonstrates superior performance, improving 1.0\% in Dice and 2.0\% in PQ over the entropy-based strategy. Additionally, it shows a gain of 1.1\% in Dice and 2.1\% in PQ when compared to the threshold-based method. All pseudo-labeling strategies were evaluated under the same settings to ensure a fair comparison.

\begin{table}[t]
\centering
\caption{Performance Comparison using  different pseudo-label selection strategies for the segmentation on EPFL $\rightarrow$R.}
\setlength{\tabcolsep}{9mm}
\label{tab:3}
\begin{tabular}{lcc}
\toprule
Pseudo-Labeling for Segmentation                  & Dice (\%)       & PQ (\%)                        \\ \midrule
Threshold-based strategy    &  \multicolumn{1}{c}{91.5}             &   \multicolumn{1}{c}{73.5}       \\
Entropy-based strategy \citep{qiu2024weakly}           &    \multicolumn{1}{c}{91.6}     &   \multicolumn{1}{c}{73.6}      \\
Instance-aware based strategy (Ours)    & \multicolumn{1}{c}{92.6}                 & \multicolumn{1}{c}{75.6}                 \\ \bottomrule
\end{tabular}
\end{table}

\begin{table}[t]
\centering
\caption{Influence of different types of contrastive alignment. Segmentation results of our model (15\%) for EPFL $\rightarrow$ R are reported. While `Source $\&$ Target' refers to conducting intra-domain contrastive alignment on both domains, `Source $+$ Target' refers to conducting both intra-domain alignment on the target domain and inter-domain alignment across domains. }
\setlength{\tabcolsep}{4mm}
\label{tab:4}
\begin{tabular}{llcc}
\toprule
Contrast. Alignment &  Settings &  Class-focused Contrastive                & \multicolumn{1}{c}{Dice (\%)}                                 \\ \midrule
   No Contrast      &- &  -  &  \multicolumn{1}{c}{91.1}                    \\
\midrule
\multirow{3}{*}{Intra-domain}& Source only     &  -   &  \multicolumn{1}{c}{91.4}                  \\
     &Target only  &     - &    \multicolumn{1}{c}{91.6}                                \\
 &Source \& Target  &    -  &    \multicolumn{1}{c}{91.9}                               \\ \midrule
 \multirow{2}{*}{Inter- \&
 Intra- domain}&  Source $+$ Target\tnote{\dag}                & Background   &     \multicolumn{1}{c}{91.9}                               \\
 & Source   $+$ Target\tnote{\dag}             &   Foreground     &     \multicolumn{1}{c}{92.6}                               \\ \bottomrule
\end{tabular}
\footnotetext{$\dag$ Utilizing only target quires in the contrastive alignment.}
\end{table}

\begin{figure}[t]
\centering
\includegraphics[width=1\textwidth]{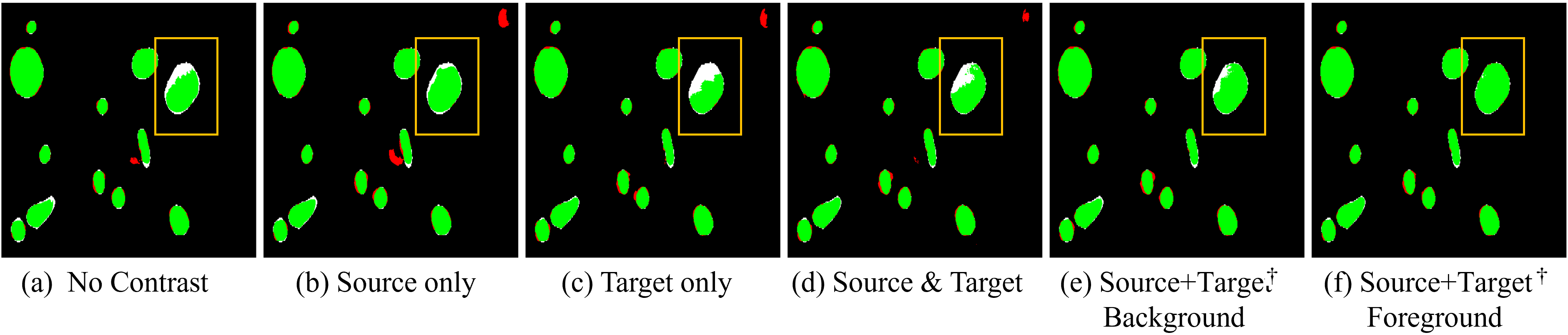}
\caption{Visualization of segmentation using different types of contrastive alignment.  Green: true positives; Red: false positives; White: false negatives.
  } \label{fig:5}
\end{figure}

\textbf{Analysis of the class-focused contrastive alignment}.
In our model, we  conduct both intra-domain contrastive learning on the target domain and inter-domain contrastive alignment with a focus on the foreground class within a single contrastive loss.  More specifically, the foreground prototype $\bar{\mu}^1$  fuses both source and target prototypes. Table \ref{tab:4}  compares our approach to five alternative settings: 1)  no contrastive learning; 2) only intra-domain contrastive learning on the source domain; 3) only intra-domain contrastive learning on the target domain; 4) intra-domain contrastive learning on both the source and target domains with two separate losses; 5) conducting both intra-  and inter-domain contrastive alignment with a focus on the background class within a single contrastive loss. Notably, using two separate losses, in contrast to utilizing a single loss as our model, leads to an inevitable increase in computational burden.
 The results indicate that our method outperforms all five alternative settings. Additionally, the findings demonstrate that the source prototype is informative and positively contributes to our class-focused contrastive alignment. The visual comparison results in Fig. \ref{fig:5} further confirm the effectiveness of conducting class-focus inter-domain alignment and the combination of inter- and intra- domain alignment.

\textbf{Influence of different amounts of point annotations}.
Table \ref{tab:5} highlights a key comparison of two aspects. First, it reveals that as annotation level increases, our model shows improved results. When supervised with 100\% of the center points, our model achieves a Dice score of  93.4\%, which is just 0.2\% lower than  the score of the fully supervised oracle, as presented in Table \ref{tab:1}. However, annotating all center points remains time-consuming and requires extensive expert knowledge \citep{qiu2024weakly}.  In contrast, our model,  utilizing only  15\% of the center points as the annotation, strikes a good balance of segmentation performance and annotation efficiency. Second, our model consistently outperforms the WDA-Net across varying amounts of target annotations.
\begin{table}[t]
\caption{Performance comparison on EPFL Hippocampus$\rightarrow$MitoEM-R with different amounts of target annotations, measured in Dice (\%) performance. }
\centering
\setlength{\tabcolsep}{4mm}
\label{tab:5}
\begin{tabular}{lcccccc}
\toprule
Method&0\% & 5\%&15\% &50\% & 100\% & Oracle\\
\midrule
WDA-Net \citep{qiu2024weakly}& 86.1 & 88.8& 90.1&91.4 &92.0& 93.6\\
Ours& 90.2& 91.8& 92.6 &92.7 & 93.4&93.6\\
\bottomrule
\end{tabular}
\end{table}

  \begin{figure}[t]
\centering
\includegraphics[width=0.9\textwidth]{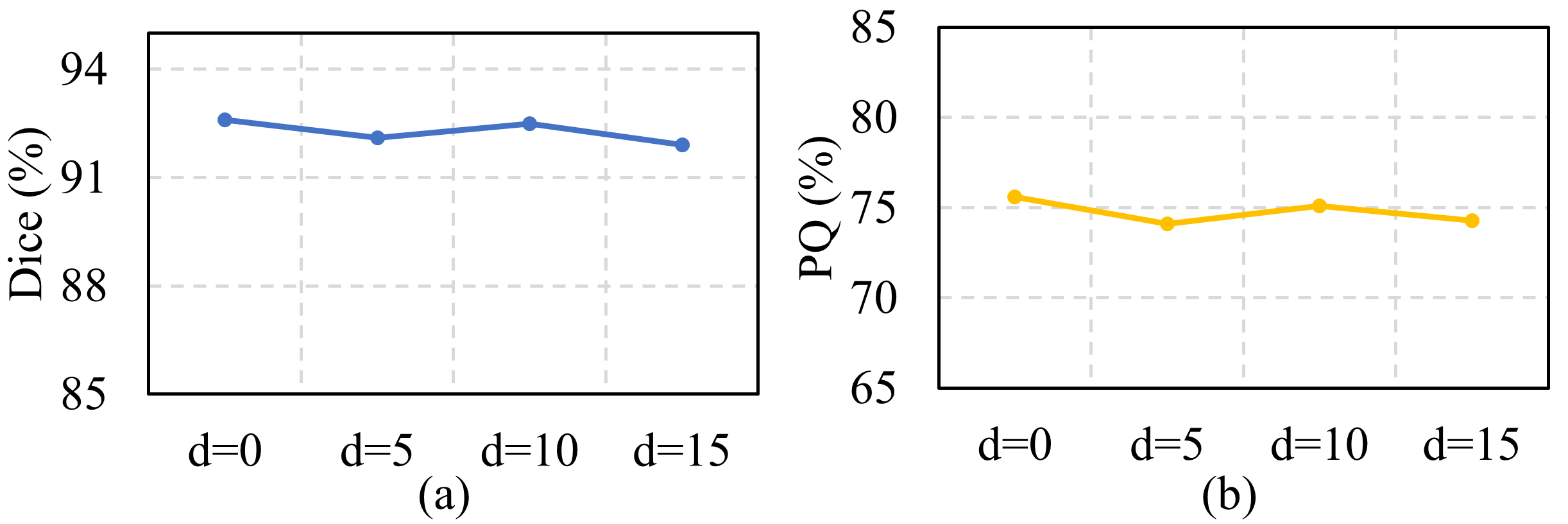}
\caption{Robustness to location deviations of the sparse point annotation.
The performance of our model (15\%) on EPFL Hippocampus$\rightarrow$MitoEM-R is evaluated when randomly displacing the sparse point annotations by $d$ pixels from their respective centers.
  } \label{fig:6}
\end{figure}

\textbf{Robustness to sample deviations}.
In Fig. \ref{fig:6}, we investigate the impact of the location deviation of the point annotations, i.e., annotation noise. Specifically, we achieved this by randomly displacing the original point annotations by $d$ pixels from their respective centers, while ensuring that all deviations remain within the foreground object instances.
 The comparative  results in Fig. \ref{fig:6} indicate that our model (15\%) shows robust resilience to annotation noise, which is desirable for facilitating efficient annotation processes.

 In Table \ref{tab:6}, we further examine the influence of sampling deviation on our model's performance. Specifically, we trained our model (15\%) with multiple samples and assessed its performance accordingly. The comparative results demonstrate that the performance of our model shows a standard deviation of 0.1 in Dice and 0.3 in PQ, highlighting the robustness of our model against sampling variations.

 \textbf{Influence of the temperature parameter $\tau$}.
 Table \ref{tab:7} illustrates the influence of the temperature parameter $\tau$ in the contrastive alignment loss $\mathcal{L}_{contrast}$, where $\tau$ controls the softness of the softmax function and also influences the penalty strength on hard negative samples in the contrastive loss \citep{wang2021understanding}. Notably, $\tau$=1 refers to the standard softmax function, while  $\tau<$1 means a more sharpening function and $\tau>$1 means a softer function.  Table \ref{tab:7} shows that the proposed model obtains the highest performance when the temperature parameter $\tau$ is 0.1. For our default setting $\tau$=0.5, the model also shows relatively higher performance in Dice.

\begin{table}[t]
\centering
\setlength{\tabcolsep}{3mm}
\caption{Influence of the sample deviations of the sparse point annotation. The performance of our model (15\%) on EPFL Hippocampus$\rightarrow$MitoEM-R with different sampling was evaluated. }
\label{tab:6}
\renewcommand\arraystretch{1.5}
\begin{tabular}{lcccccc}
\toprule
&Sample 1& Sample 2&Sample 3&Sample 4&Sample 5&Mean$\pm$Std \\
\midrule
Dice (\%)& 92.6& 92.6&92.4&92.5&92.4&\multicolumn{1}{c}{92.5$\pm$0.1}\\
PQ (\%)&76.1&75.6&75.5&75.6&75.6&75.8$\pm$0.3\\
\bottomrule
\end{tabular}
\end{table}

\begin{table}[t]
\caption{Influence of the temperature parameter $\tau$. }
\centering
\setlength{\tabcolsep}{6mm}
\label{tab:7}
\begin{tabular}{lcccccc }
\toprule
 & 0.05&0.1 &0.5& 0.7 &1.0 &1.5\\
\midrule
Dice (\%)    &92.5 &92.8 &92.6 &92.5  &92.5 &92.3\\
PQ (\%) &75.3 &76.5 &75.6 &75.5  &75.9 &75.8\\
\bottomrule
\end{tabular}
\end{table}
\section{Discussion and Conclusion}
This study focuses on segmenting mitochondria from nano-scale electron microscopy (EM) images using a WDA-based approach. Quantitative analysis of subcellular organelles, particularly mitochondria, plays critical roles in cellular homeostasis and various diseases, including cardiovascular diseases, neurodegenerative disorders, and cancers \citep{alston2017genetics}. Quantitative information about the morphology, distribution, and number is crucial for the development of therapeutic strategies \citep{neikirk2023call}.

To effectively segment numerous mitochondria from EM images, we propose an approach that integrates instance-aware pseudo-labeling and class-focused contrastive alignment within a multitask learning framework. Our method includes an auxiliary center detection mechanism to enhance the quality of the segmentation pseudo-labels. We construct foreground-class-focused prototypes, which employ the prototypes of the source foreground class to enhance the learning of latent representations. By using just a few center points as additional weak labels, our method outperforms SOTA WDA and UDA methods as well as SAM-like methods in terms of both class-level and instance-level metrics by a large margin. Moreover, we achieve performance close to the supervised upper bound with minimal annotation cost, which is essential for the quantitative analysis of various types of EM images from different species in real-world scenarios.

Despite the greatly reduced annotation cost on the target domain, our method has several limitations. First, inconsistencies in segmentation occur between slices due to the use of a 2D segmentation strategy.  In contrast,  2.5D approaches show greater consistency across slices \citep{huang2022domain}.  Second, our model relies on source data and the corresponding dense labels for model training, which is a limitation in real-world scenarios with privacy concerns. In further work, we will investigate the WDA under the source-free setting.

\backmatter

\section*{Declarations}
\bmhead{Funding}
This work was partly supported by the National Natural Science Foundation of China under Grant 12471498 and the Natural Science Foundation of Xiamen under Grant 3502Z202373042.

\bmhead{Conflict of interest}
The authors declare no competing interests.

\bmhead{Author contribution}
All authors contributed to the overall conception of the study. Shan Xiong was responsible for data preprocessing, initial analysis, model development, and software implementation. Jiabiao Chen contributed to methodology refinement and experimental validation. Shan Xiong, Ye Wang, and Jialin Peng wrote the main manuscript
text, with contributions from all authors. Ye Wang and Jialin Peng designed the research framework and contributed to the investigation and
supervision, and were involved in the critical reviewing and editing of
the manuscript. All authors read and approved the final manuscript.


\end{document}